\documentclass[journal]{IEEEtran}
\usepackage{times}
\usepackage{array}
\usepackage{epsfig}
\usepackage{graphicx}
\usepackage{amsmath}
\usepackage{amssymb}
\usepackage{mathtools}
\usepackage{colortbl}

\usepackage{CJKutf8}

\usepackage{float}
\usepackage{color}
\usepackage{subfig}
\usepackage{multirow}
\usepackage{booktabs}
\usepackage{epstopdf}

\usepackage{algorithm}
\usepackage{algpseudocode}
\ifCLASSINFOpdf
\else
\fi
\begin{document}
\begin{CJK}{UTF8}{gbsn}
\captionsetup[figure]{labelfont={bf},name={Fig.},labelsep=period}
%
\title{Lexicon-constrained Copying Network for Chinese Abstractive Summarization}
%
%
%
\author{Boyan~Wan,Mishal Sohail
\thanks{Boyan~Wan is with the College of Computer Science and Electronic Engineering, Hunan University, and the National Supercomputing Center in Changsha, Hunan, Changsha, 410082, China(e-mail:  wanboyan@hnu.edu.cn)}
\thanks{Department of Industrial Engineering, International Ataturk Alatoo University, Bishkek, Kyrgyzstan.}
}
\maketitle

\begin{abstract}
Copy mechanism allows sequence-to-sequence models to choose words from the input and put them directly into the output, which is finding increasing use in abstractive summarization. However, since there is no explicit delimiter in Chinese sentences, most existing models for Chinese abstractive summarization can only perform character copy, resulting in inefficient. To solve this problem, we propose a lexicon-constrained copying network that models multi-granularity in both encoder and decoder. On the source side, words and characters are aggregated into the same input memory using a Transformer-based encoder. On the target side, the decoder can copy either a character or a multi-character word at each time step, and the decoding process is guided by a word-enhanced search algorithm which facilitates the parallel computation and encourages the model to copy more words. Moreover, we adopt a word selector to integrate keyword information. Experiments results on a Chinese social media dataset show that our model can work standalone or with the word selector. Both forms can outperform previous character-based models and achieve competitive performances.
\end{abstract}


%
\IEEEpeerreviewmaketitle

\section{Introduction}
In recent years, abstractive summarization \cite{Abstractive} has made impressive progress with the development of sequence-to-sequence (seq2seq) framework \cite{seq2seq1,seq2seq2}. This framework is composed by an encoder and a decoder. The encoder processes the source text and extracts the necessary information for the decoder, which then predicts each word in the summary.
Thanks to their generative nature, abstractive summaries can include novel expressions never seen in the source text, but at the same time, abstractive summaries are more difficult to produce compared with extractive summaries \cite{extractive1,extractive2} which formed by directly selecting a subset of the source text.
It has been also found that seq2seq-based abstractive methods usually struggle to generate out-of-vocabulary (OOV) words or rare words, even if those words can be found in the source text. Copy mechanism \cite{pointer0} can alleviate this problem and meanwhile maintain the expressive power of the seq2seq framework. The idea is to allow the decoder not only to generate a summary from scratch but also copy words from the source text.

Though effective in English text summarization, the copy mechanism remains relatively undeveloped in the summarization of some East Asian languages e.g. Chinese. Generally speaking, abstractive methods for Chinese text summarization comes in two varieties, being word-based and character-based. Since there is no explicit delimiter in Chinese sentence to indicate word boundary, the first step of word-based methods \cite{LCSTS} is to perform word segmentation \cite{seg1,seg2}. Actually, in order to avoid the segmentation error and to reduce the size of vocabulary, most of the existing methods are character-based \cite{global,vae1,contrastive}. When trying to combine the character-based methods in Chinese with copy mechanism, the original ``word copy'' degrades to ``character copy'' which does not guarantee a multi-character word to be copied verbatim from the source text \cite{pointer1}. Unfortunately, copying multi-character words is quite common in Chinese summarization tasks. Take the Large Scale Chinese Social Media Text Summarization Dataset (LCSTS) \cite{LCSTS} as an example, according to Table I, about 37\% of the words in the summaries are copied from the source texts and consist of multiple characters. 

\begin{table}  
	\caption{Percentage of different types of words occur in the summaries of Chinese text summarization trainning data. }  
	\begin{center}  
		\resizebox{0.7\linewidth}{!}{
			\begin{tabular}{ccc}  
				\hline
				\textbf{Word Len.}&\textbf{Copied}&\textbf{Generated}\\ \hline
				1&21.6\%&12.3\%\\
				2&28.9\%&21.8\%\\
				$\geqslant$3&7.6\%&7.7\%\\\hline
			\end{tabular}
		}
	\end{center}  	  
	\label{weibo}
\end{table}

Selective read \cite{pointer1} was proposed to handle this problem. It calculates the weighted sum of encoder states corresponding to the last generated character and adds this result to the input of the next decoding step. Selective read can provide location information of the source text for the decoder and help it to perform the consecutive copy. A disadvantage of this approach, however, is that it increases reliance of present computation on partial results before the current step which makes the model more vulnerable to the errors accumulation and leads to exposure bias\cite{exposure} during inference. 
Another way to make copied content consecutive is through directly copying text spans. Zhou et al. \cite{seqcopynet} implement span copy operation by equipping the decoder with a module that predicts the start and end positions of the span. Because a longer span can be decomposed to shorter ones, there are actually many different paths to generate the same summary during inference, but their model is optimized by only the longest common span at each time step during training, which exacerbates the discrepancy between two phases.
In this work, we propose a novel lexicon-constrained copying network (LCN). The decoder of LCN can copy either a single character or a text span at a time, and we constrain the text span to match a potential multi-character word. Specifically, given a text and several off-the-shell word segmentators, if a text span is included in any segmentation result of the text, we consider it as a potential word.
By doing so, the number of available spans is significantly reduced, making it is viable to marginalize over all possible paths during training. Furthermore, during inference, we aggregate all partial paths on the fly that producing the same output using a word-enhanced beam search algorithm, which encourages the model to copy multi-character words and facilitates the parallel computation.

To be in line with the aforementioned decoder, the encoder should be revised to learn the representations of not only characters but also multi-character words. In the context of neural machine translation, Su et al. \cite{lattice1} first organized characters and multi-character words in a directed graph named word-lattice. Following Xiao et al. \cite{lattice2}, we adopt an encoder based on the Transformer \cite{transformer} to take the word-lattice as input and allow each character and word to have its own hidden representation. By taking into account relative positional information when calculating self-attention, our encoder can capture both global and local dependencies among tokens, providing an informative representation of source text for the decoder to make copy decisions. 

Although our model is character-based (because only characters are included in the input vocabulary), it can directly utilize word-level prior knowledge, such as keywords. In our setting, keywords refer to words in the source text that have a high probability of inclusion in the summary. Inspired by Gehrmann et al. \cite{bottom}, we adopt a separate word selector based on the large pre-trained language model, e.g. BERT \cite{Bert} to extract keywords. When the decoder intends to copy words from the source text, those selected keywords will be treated as candidates, and other words will be masked out.  Experimental results show that our model can achieve better performance when incorporating with the word selector.

\section{Related Work}
Most existing neural methods to abstractive summarization fall into the sequence to sequence framework. Among them, models based on recurrent neural networks (RNNs) \cite{RNN1,RNN2,RNN3} are more common than those built on convolutional neural network (CNNs) \cite{Abstractive,cnn2}, because the former models can more effectively handle long sequences. Attention \cite{atten} is easily integrated with RNNs and CNNs, as it allows the model to focus more on salient parts of the source text \cite{atten2,atten3}. Also, it can serve as a pointer to select words in the source text for copying \cite{pointer0,pointer1}. In particular, architectures that are constructed entirely of attention, e.g. Transformer \cite{transformer} can be adopted to capture global dependencies between source text and summary \cite{bottom}.  

Prior knowledge has proven helpful for generating informative and readable summaries. Templates that are retrieved from training data can guide summarization process at the sentence-level when encoded in conjunction with the source text \cite{template1,template2}. Song et al. \cite{structure} show that the syntactic structure can help to locate the content that is worth keeping in the summary, such as the main verb. 
Keywords are commonly used in Chinese text summarization.
When the decoder is querying from the source representation, Wang and Ren \cite{keyword1} use the keywords extracted by the unsupervised method to exclude noisy and redundant information.
Deng et al. \cite{keyword2} propose a word-based model that not only utilizes keywords in the decoding process, but also adds the keywords produced by the generative method into the vocabulary in the hope of alleviating out of vocabulary (OOV) problem. Our model is drastically different from the above two models in terms of the way keywords being extracted and encoded.

The most related works are in the field of neural machine translation, in which many researchers resort to the assistance of multi-granularity information.  
On the source side, Su et al. \cite{lattice1} use an RNN-based network to encode the word-lattice, an input graph that contains both word and character. Xiao et al. \cite{lattice2} apply the lattice-structured input to the Transformer \cite{transformer} and generalize the lattice to construct at a subword level.
To fully take advantage of multi-head attention in the Transformer, Nguyen et al.\cite{granularity1} first partition input sequence to phrase fragments based on n-gram type and then allow each head to attend to either one certain n-gram type or all different n-gram types at the same time.
In addition to n-gram phrases, the multi-granularity self-attention proposed by Hao et al. \cite{granularity2} also attends to syntactic phrases obtained from syntactic trees to enhance structure modeling. 
On the target side, when the decoder produces an UNK symbol which denotes a rare or unknown word, Luong et al. restore it to a natural word using a character-level component. 
Srinivasan et al. \cite{subword2} adopt multiple decoders that map the same input into translations at different subword-levels, and combine all the translations into the final result, trying to improve the flexibility of the model without losing semantic information.
While our model and the above models all utilize multi-granularity information, our model differs at that we impose a lexical constraint on both encoding and decoding.

\begin{figure*}
	\centering
	\includegraphics[width=0.8\linewidth]{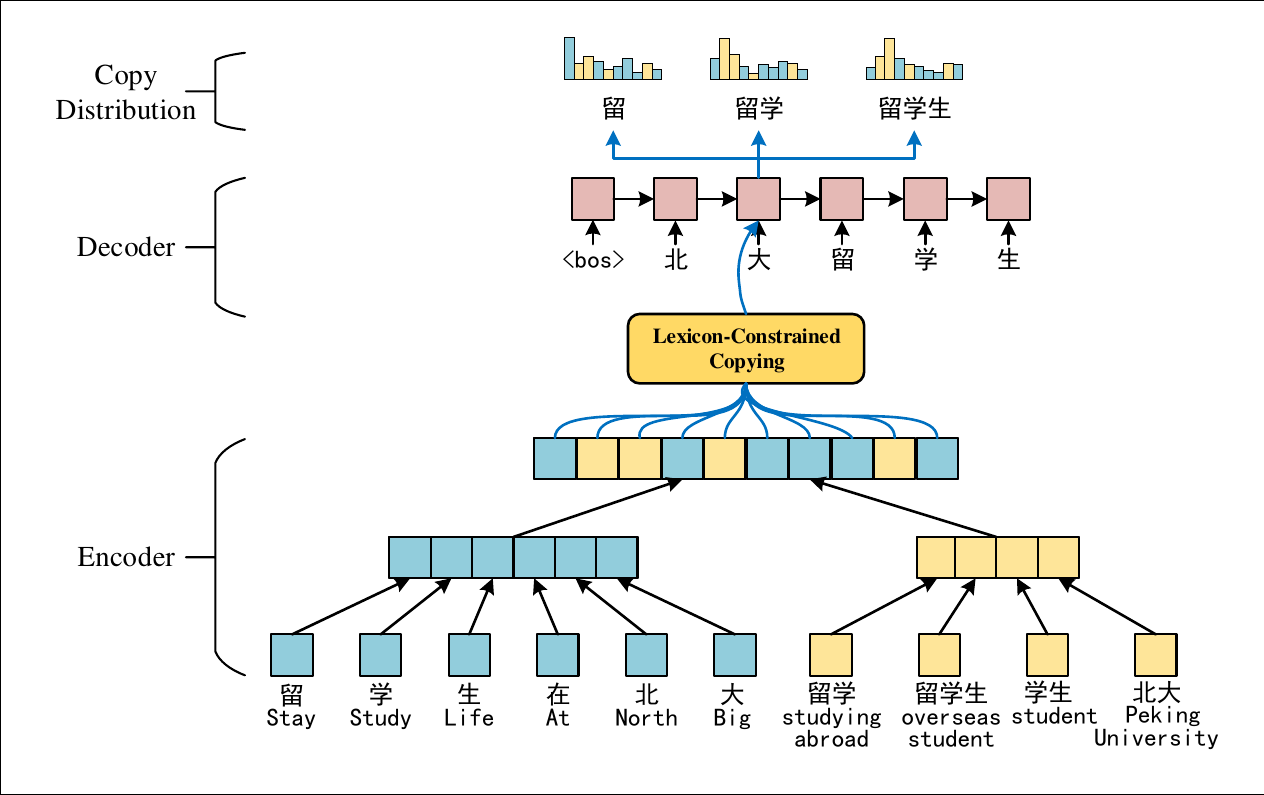}
	\caption{Structure overview of the proposed model.}
	\label{fig:first}
\end{figure*}

\section{Model}
This section describes our proposed model in detail.
\subsection{Notation}
Let character sequence $x_{1:I}=\{x_1,...,x_I\}$ be a source text, 
we can define a text span $x_{i:j}$ that starts with $x_i$ and ends with $x_j$ a potential word if it is contained by any word segmentation result of $x_{1:I}$. Because both characters and words can be regarded as tokens, we include all characters and potential words of the source text in a token sequence $o_{1:M}=\{o_1,...,o_M\}$.

\subsection{Input Representation }
\label{sec:input} 
Given a token $o_m=\{x^1,...,x^l\}$, where $l$ is the token length ($l=1$ when $o_m$ is a character), we first convert it into a sequence of vectors, using the character embedding $\textbf{E}^c$. Then a bi-directional Long Short-term Memory Network (bi-LSTM) is applied to model the token composition:
\begin{equation}
\textbf{g}_m=[\overleftarrow{LSTM}(\textbf{E}^c(x^{1})),\overrightarrow{LSTM}(\textbf{E}^c(x^{l}))]
\end{equation}
Where $\textbf{g}_m\in\mathbb{R}^{d}$ denotes the input token representation, which is formed by concatenating the backward state of the beginning character and the forward state of the ending character. 

Since the Transformer has no sequential structure, Vaswani et al. \cite{transformer} proposed positional encoding to explicitly model the order of the sequence. In this work, we assign each token an absolute position which depends on the first character of the token. For example, the absolute position of the word ``留学 (studying abroad)'' in Fig.~\ref{fig:first} is the same as that of the character ``留 (stay)''. By adding the encoding of the absolute position to the token representation, we can get the final input representation $\textbf{G}=\{\textbf{g}_1,...,\textbf{g}_M\}$. 
\subsection{Encoder} 
\begin{table}
	\newcommand{\tabincell}[2]{\begin{tabular}{@{}#1@{}}#2\end{tabular}}
	\scriptsize
	\caption{Position of $x_{c:d}$ relative to $x_{a:b}$ under different conditions. Constant $r$ is used to limit the relative position from $0$ to $2r+3$, where $2r+1$ to $2r+3$ each represent special cases: $x_{c:d}$ includes $x_{a:b}$, $x_{c:d}$ is included in $x_{a:b}$, and $x_{c:d}$ intersects with $x_{a:b}$. }
	\begin{center}
		\resizebox{0.7\linewidth}{!}{
			\begin{tabular}{|c|c  |}
				\hline
				\textbf{Conditions}&\textbf{Relative Position}\\ \hline	
				$d< a$ &$max(0,r-a+d)$\\ \hline
				$a=c$ and $b=d$ & $r$\\ \hline
				$b< c$ &$min(2r,r+c-b)$\\ \hline
				\tabincell{c}{$c \leq\ a \leq b<d$ or \\ $c<a \leq b \leq d$} &$2r+1$\\ \hline
				\tabincell{c}{$a \leq c \leq d<b$ or \\ $a<c \leq d \leq b$} &$2r+2$\\ \hline
				
				otherwise &$2r+3$\\ \hline
			\end{tabular}
		}
	\end{center}  	
	\label{relations}
\end{table}
However, absolute position alone cannot precisely reflect the relationship among tokens. Consider again the example in Fig.~\ref{fig:first}, the distance between the word ``留学 (studying abroad)'' and the character ``生 (life)'' is 2 according to their absolute positions, but they are actually neighboring tokens in a certain segmentation. To alleviate this problem, Xiao et al. \cite{lattice2} extend the Transformer \cite{transformer} by taking into account relation types when calculating self-attention. In this work, we adopt relative position as an alternative to relation type. The main idea is that relative position is complementary to absolute position and can guide each token to interact with other tokens in a coherent manner.
Given two tokens $x_{a:b}$ and $x_{c:d}$ that correspond to a span of the source text each, the position of $x_{c:d}$ relative to $x_{a:b}$ is determined by both their beginning and ending characters as shown in Table \ref{relations}. Following Xiao et al. \cite{lattice2}, we revise self-attention to integrate relative positional information. Concretely, a self-attention layer consists of $h$ heads, which operate in parallel on a sequence $\textbf{H}=\{\textbf{h}_1,...,\textbf{h}_M\}$ of context vectors with dimension $d$. After modification, the resulting output \textbf{attn} for each attention head is defined as:
\begin{equation}
e_{ij}=\frac{(\textbf{W}^q\textbf{h}_i ) (\textbf{W}^k \textbf{h}_j + \textbf{p}_{ij}^k)^T}{\sqrt{d_z}}
\end{equation}
\begin{equation}
a_{ij}=\frac{exp(e_{ij})}{\sum_{k=1}^{M}exp(e_{ik})}
\end{equation}
\begin{equation}
\textbf{attn}_{i}=\sum_{j=1}^{M}a_{ij}(\textbf{W}^v \textbf{h}_j+ \textbf{p}_{ij}^v)
\end{equation}
where $\textbf{W}^q$, $\textbf{W}^k$, $\textbf{W}^v \in\mathbb{R}^{d_z \times d}$ are all model parameters, $d_z=d/h$ is the hidden dimension for each head, $\textbf{p}_{ij}^k$ and $\textbf{p}_{ij}^v$ are learned embeddings that encode the position of token $t_j$ relative to token $t_i$. We concatenate the outputs of all heads to restore their dimension to $d$, and then apply other sub-layers (such as feed-forward layer) used in the original Transformer \cite{transformer} to get the final output of the layer. Several identical self-attention layers are stacked to build our encoder. For the first layer, $\textbf{H}$ is input representation $\textbf{G}$. For the subsequent layers, $\textbf{H}$ is the output of the previous layer.

\subsection{Decoder}
The encoder proposed by Xiao et al. \cite{lattice2} takes both characters and words as input and thus has the ability to learn multi-granularity representations. However, as their decoder is character-based which consumes and outputs only characters, the word representations induced from the encoder cannot receive supervision signal directly from the decoder and remain a subsidiary part of the input memory. To alleviate this problem, we extend the standard Transformer decoder by a lexicon-constrained copying module, which not only allows the decoder to perform multi-character word copy but also provides auxiliary supervision on word representations. Specifically, at each time step $t$ we leverage a single-head attention over the input memory $\textbf{H}=\{\textbf{h}_1,...,\textbf{h}_M\}$ and the decoder hidden state $\textbf{s}_t$ to produce copy distribution $\textbf{a}_t$ and context vector $\textbf{c}_t$:
\begin{equation}
e_{tj}=\frac{(\textbf{W}_{copy}^q \textbf{s}_t ) (\textbf{W}_{copy}^k \textbf{h}_j )^T}{\sqrt{d}}
\end{equation}
\begin{equation}
\label{mask}
a_{tj}=\frac{exp(e_{tj})}{\sum_{k=1}^{M}exp(e_{tk})}
\end{equation}
\begin{equation}
\textbf{c}_{t}=\sum_{j=1}^{M}a_{tj}(\textbf{W}_{copy}^v \textbf{h}_j)
\end{equation}
In addition to the predefined character vocabulary $\mathcal V$ and a special token UNK denoting any out-of-vocabulary token, lexicon-constrained copying module expand the output space by two sets $\mathcal C$ and $\mathcal W$, consisting of characters and multi-character words that appear in the source text, respectively. So that the probability of emitting any token $o$ at time step $t$ is:
\begin{equation}
P(o|\cdot)=\left\{
\begin{array}{lc}
p_{gen}gen(o) 
+ p_{copy}\sum_{i:o_i=o}a_{ti}
& o \in \mathcal V \cup \mathcal C \\
p_{copy}\sum_{i:o_i=o}a_{ti}           & o \in \mathcal W\\
p_{gen}gen(\text{UNK} )        & otherwise \\
\end{array} \right.
\end{equation}
where $p_{gen}$, $p_{copy}\in [ 0,1 ]$ control the decoder switching between generation-mode and copy-mode, $gen(\cdot)$ provide a probability distribution over character vocabulary $\mathcal V$:
\begin{equation}
p_{gen}=\sigma(\textbf{V}_g[\textbf{s}_t,\textbf{c}_t]+\textbf{b}_p)
\end{equation}
\begin{equation}
p_{copy}=1-p_{gen}
\end{equation}
\begin{equation}
gen(\cdot)=\text{softmax}(\hat{\textbf{V}_g} (\textbf{V}_g[\textbf{s}_t,\textbf{c}_t]+\textbf{b})+\hat{\textbf{b}_g})
\end{equation}
where $\sigma(\cdot)$ is the sigmoid function.

With the introduction of lexicon-constrained copying module, our decoder can predict tokens of variable lengths at each time step, and thereby can generate any segmentation of a sentence. Naturally, we expect to evaluate the probability of a summary by marginalizing over all its segmentations.  For example, the probability of a summary consisting of only one word ``北京 (BeiJing)'' can be factorized as:
\begin{equation}
\begin{aligned}
&P(\text{北,京})=P(\text{北}|\epsilon)P(\text{京}|\text{北})P(\epsilon|\text{北,京})\\
&\qquad+P(\text{北京}|\epsilon) P(\epsilon|\text{北京}) \nonumber
\end{aligned}
\end{equation}
where each term corresponds to a segmentation and is the product of conditional probabilities, we use $\epsilon$ to denote either the beginning or end of a sentence. Note that the conditional probability here depends on the current segmentation, which means the decoder directly take tokens in a segmentation as input. However, if we feed the decoder with character-level input and reformulate the conditional probability, the above probability can be rewritten as:
\begin{equation}
\!P(\text{北,京})=P(\epsilon|\text{北,京})\Big(P(\text{北}|\epsilon)  P(\text{京}|\text{北})\!
+P(\text{北京}|\epsilon)\Big)\\ \nonumber
\end{equation}
where we factor out $P(\epsilon|\text{北,京})$, because it is shared by two different segmentations. As can be seen from the above example, the assumption that conditional probability of a token depends only on its preceding character sequence facilitate the reuse of computation and thus makes it feasible to apply dynamic programming. Formally, let character sequence $y_{1:J}=\{y_1,...,y_J\}$ be a summary, its probability can be represented as a recursion:
\begin{equation}
\label{recur}
P(y_{1:J})=\quad\sum_{\mathclap{\tiny\begin{array}{c} o\in \mathcal V \cup \mathcal C \cup \mathcal W \\   \!o=y_{J-\ell+1:J}\! \end{array}}}\quad P(o|y_{1:J-\ell})P(y_{1:J-\ell})
\end{equation}
Note that all the above $P(\cdot)$ are inevitably conditioned on the source text, so we omit it for simplicity. We train the model by maximizing $P(y_{1:J})$ for all training pair in the dataset.

During inference, since there is no access to the ground truth, we need a search algorithm which can guide the generation of the summary in a left-to-right fashion. Beam search \cite{beam} is the most common search algorithm in seq2seq frameworks, but cannot be used directly in our scenario. To illustrate this, we first define hypothesis as a partial output that consists of tokens. Hypotheses can be further divided into character hypotheses and word hypotheses based on whether their last token is a character or a multi-character word. For hypotheses within a beam, the standard beam search algorithm updates states for them by feeding their last tokens to the decoder and then generate new hypotheses through suffixing them with a token sampled from the model's distribution. Because our decoder is designed to take only characters as input, multiple decoder steps are required to update the state for a word hypothesis. As a result, it is difficult to conduct batched update for a beam containing both word hypotheses and character hypotheses. To this end, we proposed a novel word-enhanced beam search algorithm, where the beam is also split into two parts: the character beam and the word beam. 
The word beam is used to update the states for word hypotheses. When their states are fully updated, word hypotheses are placed into the character beam (see lines 5-8 of Algorithm~\ref{alg:beam}). Note that we do not perform generation step for word hypotheses in the word beam, that is to say, with the same length, the more multi-character words a hypothesis includes, the more generation steps it can skips, which may give it a higher probability.
\begin{algorithm}[htp]
	
	\caption{Pseudo-code for word-enhanced Beam Search}
	\label{alg:beam}  
	\begin{algorithmic}[1]
		
		\Require 
		$model$, $source$, maximum summary length $L$, beam size $k$
		
		\State $startHyp$$\leftarrow$getStartHyp($\epsilon$)
		\State $B_c$=\{$startHyp$\},$B_w$=\{\} and $B_f$=\{\}
		\Statex $//$ Respectively the character beam, word beam, and the set of finished hypotheses
		\For{$t=0;t++;t < L$}
		\State $n,m=\{\}$
		\State $B_c,B_w$$\leftarrow$$model$.batchedUpadte($B_c,B_w,source$)
		\Statex \quad \,\,// Batched update  for hypos in both $B_c$ and $B_w$
		\For{$hyp \in B_w$}
		\If{$hyp$.isUpdated}
		\State move $hyp$ into $B_c$
		\EndIf
		\EndFor
		\State Merge hypos of the same character sequence in $B_c$
		\State $n \leftarrow$ $model$.generate($B_c$)
		\Statex \quad \,\,$//$ Generating new hypotheses and their respective 
		\Statex \quad \, log probabilities from $B_c$
		\For{$hyp \in n$}
		\If{$hyp$ ends with a multi-character word}
		\State Move $hyp$ from $n$ into $m$
		\EndIf
		\EndFor
		\State $B_w \leftarrow  \mathop{\text{k-argmax}}\limits_{hyp\in m}$ $hyp$.avgLogProb
		\State $B_c \leftarrow  \mathop{\text{k-argmax}}\limits_{hyp\in n}$ $hyp$.avgLogProb
		\For{$hyp \in B_c$}
		\If{$hyp$ ends with $\epsilon$ {\textbf{or}} $hyp$.len=$L$}
		\State Move $hyp$ from $n$ into $B_f$
		\EndIf
		\EndFor
		\EndFor
		\State $finalHyp \leftarrow \mathop{\text{argmax}}\limits_{hyp\in B_f}$ $hyp$.avgLogProb
		\State \textbf{return} $finalHyp$
		
	\end{algorithmic}
\end{algorithm}

\subsection{Word Selector}
We treat keyword selection as a binary classification task on each potential word. To obtain word representations, we first leverage BERT \cite{Bert}, a pre-trained language model to produce context-aware representations $\textbf{x}^c$ for all characters in the source text, and then feed them to a bi-LSTM network. Different from Section~\ref{sec:input} where bi-LSTM is applied to character sequence of each word, here the bi-LSTM takes the whole source character representations as input, in an attempt to build word representation that can reflect how much contextual information the word carries. Given a potential word $x_{a:b}$, where $a$ and $b$ are indexes of characters in the source text, we can calculate its final representation $\textbf{t}$ as follows:
\begin{equation}
\begin{array}{c}
\overrightarrow{\textbf{d}_a}=\overrightarrow{LSTM}(\textbf{x}^c_a),\overrightarrow{\textbf{d}_b}=\overrightarrow{LSTM}(\textbf{x}^c_b) \\
\overleftarrow{\textbf{d}_a}=\overleftarrow{LSTM}(\textbf{x}^c_a),\overleftarrow{\textbf{d}_b}=\overleftarrow{LSTM}(\textbf{x}^c_b) \\
\textbf{t}=[\overrightarrow{\textbf{d}_a},\overrightarrow{\textbf{d}_b},\overleftarrow{\textbf{d}_a},\overleftarrow{\textbf{d}_b},\overrightarrow{\textbf{d}_b}-\overrightarrow{\textbf{d}_a},\overleftarrow{\textbf{d}_a}-\overleftarrow{\textbf{d}_b}]
\end{array}
\end{equation}
then a linear transformation layer and a sigmoid function can be used sequentially on its final representation to compute the probability of $x_{a:b}$ being selected.

During training, words that appear in both summary and source text are considered as positives, rest are negatives. To make sure that decoder can access the entire source character sequence at inference time, in addition to the multi-character words with the top-$n$ probability, we treat all characters in source text as keywords. Inspired by \cite{bottom}, we utilize keyword information by masking out other words when calculating copy distribution. In particular, we leave $e_{tj}$ in Eq.~\ref{mask} unchanged for keywords and set $e_{tj}$ to zero for the rest of the words.

\section{Experiments}

\subsection{Datasets and Evaluation Metric}
We conduct experiments on the Large Scale Chinese Social Media Text Summarization Dataset (LCSTS)$\footnote{http://icrc.hitsz.edu.cn/Article/show/139.html}$ \cite{LCSTS}, which consists of source texts with no more than 140 characters, along with human-generated summaries. The dataset is divided into three parts, the (source, summary) pairs in PART II and PART III are scored manually from 1 to 5, with higher scores indicating more relevance between the source text and its summary. Following Hu et al. \cite{LCSTS}, after removing pairs with scores less than 3, PART I, PART II and PART III are used as training set, verification set and test set respectively, with 2.4M pairs in PART I, 8K pairs in PART II and 0.7K pairs in PART III.

We choose ROUGE score \cite{rouge} as our evaluation metric, which is widely used for evaluating automatically produced summaries. The metric measure the relevance between a source text and its summary based on their co-occurrence statistics. In particular, ROUGE-1 and ROUGE-2 depend on unigram and bigram overlap respectively, while ROUGE-L relies on the longest common subsequence.
\subsection{Experimental Setup}
The character vocabulary is formed by 4000 most frequent characters in the training set. To get all potential words, we use PKUSEG \cite{pkuseg}, a toolkit for multi-domain chinese word segmentation. Specifically, there are separate segmentators for four domains, including web, news, medicine, and tourism. We use these segmentators for the source text, and if a text span is included in any of word segmentation results, we regard it as a potential word.

For the lexicon-constrained copying network, we employ six attention layers of 8 heads for both encoder and decoder. Constant $r$ in TABLE~\ref{relations} is set to 8. We make character embedding and all hidden vectors the same dimension of 512 and set the filter size of feed-forward layers to 1024. For word selectors, we use a single-layer Bi-LSTM with a hidden size of 512.

During training, we update the parameters of the lexicon-constrained copying network (LCCN) and word selector by Adam optimizer with $\beta1=0.9$, $\beta2=0.98$, $\varepsilon=10^{-9}$. The same learning rate schedule of Vaswani et al. \cite{transformer} is used to the LCCN, while a fixed learning rate of 0.0003 is set for word selector. The BERT we use in the word selector is pre-trained on Chinese corpus by Wolf et al. \cite{hugging} and we freeze its parameters throughout the training.

During testing, we use a beam size of 10 and take the first 10 multi-character words predicted by the word selector and all characters in the source text as keywords.

\subsection{Baselines} 
\begin{itemize}
	\item \textbf{RNN} and \textbf{RNN-Context} are seq2seq baselines provided along with the LCSTS dataset by Hu et al. \cite{LCSTS}. Both of them have GRU encoder and GRU decoder, while RNN-context has an additional attention mechanism.
	\item  \textbf{COPYNET} integrates copying mechanism into the seq2seq framework, trying to improve both content-based addressing and location-based addressing.
	\item  \textbf{Supervison with Autoencoder (superAE)} uses an autoencoder trained on the summaries to provide auxiliary supervision for the internal representation of seq2seq. Moreover, adversarial learning is adopted to enhance this supervision.
	\item \textbf{Global Encoding} refines source representation with consideration of the global context by using a convolutional gated unit.
	\item \textbf{Keyword and Generated Word Attention (KGWA)} exploits relevant keywords and previously generated words to learn accurate source representation and to alleviate the information loss problem. 
	\item \textbf{Keyword
		Extraction and Summary Generation (KESG)} first uses a separate seq2seq model to extract keywords, and then utilize keywords information to improve the quality of the summarization. 
	\item \textbf{Transformer} and \textbf{CopyTransformer} are our implementations of the Transformer framework in the task of summarization. Copy mechanism is incorporated into \textbf{CopyTransformer}.
\end{itemize}

\subsection{Results}
\begin{table}
	\scriptsize
	\caption{Results of different models. We use $\dagger$ to indicate that the model utilizes keywords information}
	\begin{center}
		\resizebox{\linewidth}{!}{
			\begin{tabular}{l|c|c|c}
				\hline
				\textbf{Models}&\textbf{ROUGE-1}&\textbf{ROUGE-2}&\textbf{ROUGE-L}\\ \hline
				RNN \cite{LCSTS}&21.5&8.9&18.6\\
				RNN-Content \cite{LCSTS}&29.9&17.4&27.2\\
				COPYNET \cite{pointer1}&34.4&21.6&31.3\\
				superAE \cite{vae1}&39.2&26.0&36.2\\
				Global Encoding \cite{global}&39.4&26.9&36.5\\
				$\text{KESG}^{\dagger}$ \cite{keyword2} &39.4&28.4&35.3\\
				$\text{KGWA}^{\dagger}$ \cite{keyword1} &40.9&28.3&38.2\\ \hline
				Transformer&38.9&27.4&35.5\\
				CopyTransformer&39.7&28.0&35.8\\
				LCCN&41.7&29.5&38.0\\
				w/o word-enhance beam search&40.0&28.5&37.1\\
				$\text{LCCN+word selector}^{\dagger}$&\textbf{42.3}&\textbf{29.8}&\textbf{38.4}\\ \hline
			\end{tabular}
		}
	\end{center}  	
	\label{results}
\end{table}

TABLE~ \ref{results} records the results of our LCCN model and other seq2seq models on the LCSTS dataset. To begin with, we first compare two Transformer baselines. We can see that CopyTransformer outperforms vanilla Transformer by 0.8 ROUGE-1, 0.6 ROUGE-2, and 0.3 ROUGE-L, showing the importance of copy mechanism. The gap between our LCCN and vanilla Transformer is further widen to 1.8 ROUGE-1, 2.1 ROUGE-2, and 2.5 ROUGE-L, which asserts the superiority of lexicon-constrained copying over character-based copying. Compared to other latest models, our LCCN can achieve state-of-the-art performance in terms of ROUGE-1 and ROUGE-2, and is second only to the KGWA in terms of ROUGE-L. When also using keywords information as the KGWA does, LCCN+word-selector further improves the performance and overtakes the KGWA by 0.2 ROUGE-L. We also conduct an ablation study by removing the word-enhanced beam search in LCCN, denoted by w/o word-enhanced beam search in TABLE~ \ref{results}. It shows that word-enhanced beam search can boost the performance of 1.7 ROUGE-1, 1.0 ROUGE-2, and 0.9 ROUGE-L.

\subsection{Discussion}

\begin{table}
	\scriptsize
	\caption{Results of different approaches to extract keywords.}
	\begin{center}
		\resizebox{\linewidth}{!}{
			\begin{tabular}{l|c|c|c}
				\hline
				\textbf{Models}&\textbf{ROUGE-1}&\textbf{ROUGE-2}&\textbf{ROUGE-L}\\ \hline
				TFIDF &28.6&13.1&20.7\\
				encoder of LCCN&42.0&25.6&33.7\\
				word selector&46.0&28.2&36.1\\ \hline
			\end{tabular}
		}
	\end{center}  	
	\label{selector}
\end{table}
\begin{table}
	\scriptsize
	\caption{Summarization examples. Summaries in the last two blocks are separated by spaces to show the output of LCCN and LCCN+word selector per step. We use keywords to denote the multi-character words chosen by the word selector.}
	\begin{center}
		\begin{tabular}{p{8cm}}
			\hline
			\textbf{Source}: 9月5日，陕西咸阳双泉村垃圾站发现一名死亡女婴，脖子上缠有绳子。近日案件告破，令人意外的是，婴儿的父母是17岁左右的在校高中生，涉嫌勒死婴儿的，是孩子父亲。“当时看见孩子生下来心就慌了，害怕孩子哭，便用绳子勒死了”。\\
			On September 5, a dead baby girl with a rope around her neck was found at the Shuangquan Village Garbage Station in Xianyang, Shaanxi Province. Recently, the case was solved. Surprisingly, the parents of the baby were high school students around 17 years old. The father of the baby was suspected of strangulating his baby. "When I saw the baby born, I was in a panic. I was afraid of a baby crying, so I strangled her with a rope."\\ \hline
			\textbf{Reference}: 咸阳两高中生同居生下女婴因害怕孩子哭将其勒死。\\
			Two high school students in Xianyang lived together and gave birth to a baby girl and strangled her for fear of baby crying.\\ \hline
			\textbf{Transformer}: 17岁高中生当街勒死亲生父母。 \\ 
			17-year-old high school students strangled their parents in the street\\\hline
			\textbf{CopyTransformer}: 陕西17岁女婴儿缠绳子勒死。 \\
			17-year-old female infant in Shaanxi was strangled with rope\\ \hline
			\textbf{LCNN}: 17岁\quad 高中生\quad 勒\quad 死\quad 婴儿。\\
			17-year-old high school student strangled a infant\\ \hline
			\textbf{LCNN+word selector}: 高中生\quad 因\quad 害怕 \quad 勒\quad 死\quad 婴儿。\\
			High school student strangled a infant out of fear.\\ 
			\textbf{Keywords}: 女婴 (baby girl), 婴儿 (infant), 孩子 (baby), 勒死 (strangle), 害怕 (fear), 高中 (high school), 陕西 (Shaanxi), 绳子 (rope), 高中生 (high school students), 父亲 (father)\\\hline
		\end{tabular}
	\end{center}  	
	\label{example}
\end{table}
Similar to extractive summarization, we can use the top $n$ extracted keywords to form a summary, which then can be used to evaluate the quality of keywords. The first entry of TABLE~\ref{selector} shows the performance when the keywords are extracted by TF-IDF \cite{tf-idf}, a numerical statistic method that relies on the frequency of the word. The second entry shows the performance when we determine keywords based on the source representation learned by the encoder of LCCN. As can be seen from the last entry, word selector outperforms two methods mentioned above by a large margin, indicating the importance of external knowledge brought by the BERT. 

Given a source text which describes the criminal case, we
show its summaries generated by different models in TABLE~\ref{example}. It is clear that the suspect of this case is a high school student and the victim is his baby daughter. However, the summary generated by the Transformer mistakes the high school student's parents as victims and claims that the crime took place in the street, which is not mentioned in the source text. The summary of the CopyTransformer also makes a fundamental mistake, resulting the mismatch between the adjective ``17岁 (17-year-old) and the noun ``女婴儿 (female  infant)". Compared with them, the summary of our LCCN is more faithful to the source text and contains the correct suspect and victim, i.e ``高中生 (high school student)" and ``婴儿 (infant)" which are copied from the source text through only two decoder steps. With the help of word selector, our summary can further include the keyword ``害怕 (fear)" to indicate the criminal motive. 

\begin{figure}
	\centering
	\includegraphics[width=0.8\linewidth]{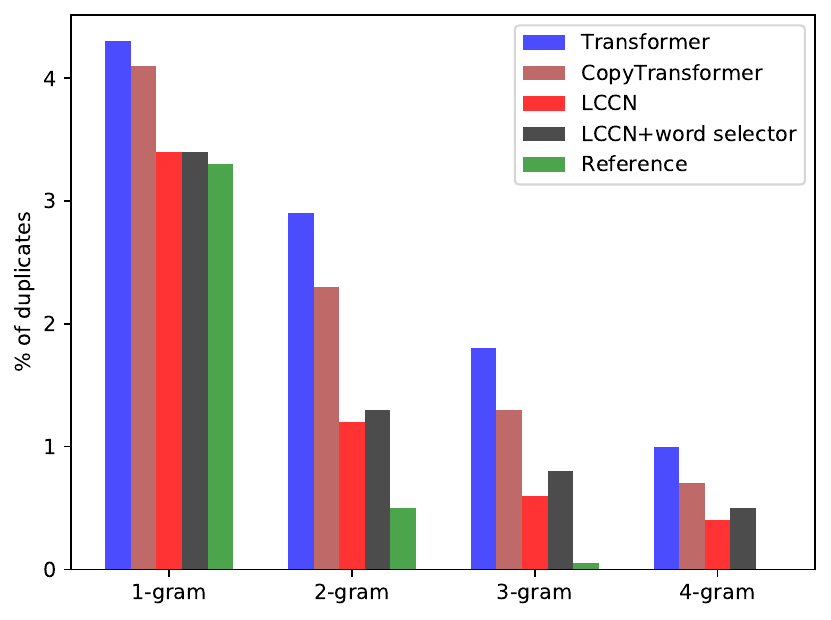}
	\caption{Percentage of the duplicates}
	\label{ngram}
\end{figure}

Compared with character-based models, our LCCN uses fewer steps to output a summary, so it should be able to reduce the possibility of repetition. To prove it, we record the percentage of n-gram duplicates for summaries generated by different models in TABLE~\ref{ngram}. The results show that our model can indeed alleviate the repetition problem, and we also notice that the repetition rate of the LCCN+word selector is slightly higher than that of LCNN, which may be due to the smaller output space after adding the word selector.

\section{Conclusion}
In this paper, we propose a novel lexicon-constrained copying network for Chinese summarization. Querying the multigranularity representation learned by our encoder, our decoder can copy either a character or a multi-character word at each time step. Experiments on the LCSTS dataset show that our model is superior to the Transformer baselines and quite competitive with the latest models. With the help of keyword information provide by the word selector, it can even achieve state-of-the-art performance. In the future, we plan to apply our model to other tasks, such as comment  generation, and to other languages, such as English.

\ifCLASSOPTIONcaptionsoff
  \newpage
\fi



%
\bibliographystyle{IEEEtran}
\bibliography{IEEEabrv,egbib.bib}

\begin{thebibliography}{10}
\providecommand{\url}[1]{#1}
\csname url@samestyle\endcsname
\providecommand{\newblock}{\relax}
\providecommand{\bibinfo}[2]{#2}
\providecommand{\BIBentrySTDinterwordspacing}{\spaceskip=0pt\relax}
\providecommand{\BIBentryALTinterwordstretchfactor}{4}
\providecommand{\BIBentryALTinterwordspacing}{\spaceskip=\fontdimen2\font plus
\BIBentryALTinterwordstretchfactor\fontdimen3\font minus
  \fontdimen4\font\relax}
\providecommand{\BIBforeignlanguage}[2]{{%
\expandafter\ifx\csname l@#1\endcsname\relax
\typeout{** WARNING: IEEEtran.bst: No hyphenation pattern has been}%
\typeout{** loaded for the language `#1'. Using the pattern for}%
\typeout{** the default language instead.}%
\else
\language=\csname l@#1\endcsname
\fi
#2}}
\providecommand{\BIBdecl}{\relax}
\BIBdecl

\bibitem{Abstractive}
A.~M. Rush, S.~Chopra, and J.~Weston, ``A neural attention model for
  abstractive sentence summarization,'' in \emph{Proceedings of the 2015
  Conference on Empirical Methods in Natural Language Processing, {EMNLP} 2015,
  Lisbon, Portugal, September 17-21, 2015}, 2015, pp. 379--389.

\bibitem{seq2seq1}
I.~Sutskever, O.~Vinyals, and Q.~V. Le, ``Sequence to sequence learning with
  neural networks,'' in \emph{Advances in Neural Information Processing Systems
  27: Annual Conference on Neural Information Processing Systems 2014, December
  8-13 2014, Montreal, Quebec, Canada}, 2014, pp. 3104--3112.

\bibitem{seq2seq2}
D.~Bahdanau, K.~Cho, and Y.~Bengio, ``Neural machine translation by jointly
  learning to align and translate,'' in \emph{3rd International Conference on
  Learning Representations, {ICLR} 2015, San Diego, CA, USA, May 7-9, 2015,
  Conference Track Proceedings}, 2015.

\bibitem{extractive1}
A.~Sharan, H.~Imran, and M.~L. Joshi, ``A trainable document summarizer using
  bayesian classifier approach,'' in \emph{First International Conference on
  Emerging Trends in Engineering and Technology, {ICETET} '08, Nagpur,
  Maharashtra, India, July 16-18, 2008}, 2008, pp. 1206--1211.

\bibitem{extractive2}
H.~Saggion and T.~Poibeau, ``Automatic text summarization: Past, present and
  future,'' in \emph{Multi-source, Multilingual Information Extraction and
  Summarization}, 2013, pp. 3--21.

\bibitem{pointer0}
C.~Gulcehre, S.~Ahn, R.~Nallapati, B.~Zhou, and Y.~Bengio, ``Pointing the
  unknown words,'' in \emph{Proceedings of the 54th Annual Meeting of the
  Association for Computational Linguistics (Volume 1: Long Papers)}, 2016.

\bibitem{LCSTS}
B.~Hu, Q.~Chen, and F.~Zhu, ``{LCSTS:} {A} large scale chinese short text
  summarization dataset,'' in \emph{Proceedings of the 2015 Conference on
  Empirical Methods in Natural Language Processing, {EMNLP} 2015, Lisbon,
  Portugal, September 17-21, 2015}, 2015, pp. 1967--1972.

\bibitem{seg1}
H.~Zhao and C.~Kit, ``Integrating unsupervised and supervised word
  segmentation: The role of goodness measures,'' \emph{Inf. Sci.}, vol. 181,
  no.~1, pp. 163--183, 2011.

\bibitem{seg2}
D.~Cai, H.~Zhao, Z.~Zhang, Y.~Xin, Y.~Wu, and F.~Huang, ``Fast and accurate
  neural word segmentation for chinese,'' in \emph{Proceedings of the 55th
  Annual Meeting of the Association for Computational Linguistics, {ACL} 2017,
  Vancouver, Canada, July 30 - August 4, Volume 2: Short Papers}, 2017, pp.
  608--615.

\bibitem{global}
J.~Lin, X.~Sun, S.~Ma, and Q.~Su, ``Global encoding for abstractive
  summarization,'' in \emph{Proceedings of the 56th Annual Meeting of the
  Association for Computational Linguistics, {ACL} 2018, Melbourne, Australia,
  July 15-20, 2018, Volume 2: Short Papers}, 2018, pp. 163--169.

\bibitem{vae1}
S.~Ma, X.~Sun, J.~Lin, and H.~Wang, ``Autoencoder as assistant supervisor:
  Improving text representation for chinese social media text summarization,''
  in \emph{Proceedings of the 56th Annual Meeting of the Association for
  Computational Linguistics, {ACL} 2018, Melbourne, Australia, July 15-20,
  2018, Volume 2: Short Papers}, 2018, pp. 725--731.

\bibitem{contrastive}
X.~Duan, H.~Yu, M.~Yin, M.~Zhang, W.~Luo, and Y.~Zhang, ``Contrastive attention
  mechanism for abstractive sentence summarization,'' in \emph{Proceedings of
  the 2019 Conference on Empirical Methods in Natural Language Processing and
  the 9th International Joint Conference on Natural Language Processing,
  {EMNLP-IJCNLP} 2019, Hong Kong, China, November 3-7, 2019}, 2019, pp.
  3042--3051.

\bibitem{pointer1}
J.~Gu, Z.~Lu, H.~Li, and V.~O.~K. Li, ``Incorporating copying mechanism in
  sequence-to-sequence learning,'' in \emph{Proceedings of the 54th Annual
  Meeting of the Association for Computational Linguistics, {ACL} 2016, August
  7-12, 2016, Berlin, Germany, Volume 1: Long Papers}, 2016.

\bibitem{exposure}
M.~Ranzato, S.~Chopra, M.~Auli, and W.~Zaremba, ``Sequence level training with
  recurrent neural networks,'' in \emph{4th International Conference on
  Learning Representations, {ICLR} 2016, San Juan, Puerto Rico, May 2-4, 2016,
  Conference Track Proceedings}, 2016.

\bibitem{seqcopynet}
Q.~Zhou, N.~Yang, F.~Wei, and M.~Zhou, ``Sequential copying networks,'' in
  \emph{Proceedings of the Thirty-Second {AAAI} Conference on Artificial
  Intelligence, (AAAI-18), the 30th innovative Applications of Artificial
  Intelligence (IAAI-18), and the 8th {AAAI} Symposium on Educational Advances
  in Artificial Intelligence (EAAI-18), New Orleans, Louisiana, USA, February
  2-7, 2018}, 2018, pp. 4987--4995.

\bibitem{lattice1}
J.~Su, Z.~Tan, D.~Xiong, R.~Ji, X.~Shi, and Y.~Liu, ``Lattice-based recurrent
  neural network encoders for neural machine translation,'' in
  \emph{Proceedings of the Thirty-First {AAAI} Conference on Artificial
  Intelligence, February 4-9, 2017, San Francisco, California, {USA}}, 2017,
  pp. 3302--3308.

\bibitem{lattice2}
F.~Xiao, J.~Li, H.~Zhao, R.~Wang, and K.~Chen, ``Lattice-based transformer
  encoder for neural machine translation,'' in \emph{Proceedings of the 57th
  Conference of the Association for Computational Linguistics, {ACL} 2019,
  Florence, Italy, July 28- August 2, 2019, Volume 1: Long Papers}, 2019, pp.
  3090--3097.

\bibitem{transformer}
A.~Vaswani, N.~Shazeer, N.~Parmar, J.~Uszkoreit, L.~Jones, A.~N. Gomez,
  L.~Kaiser, and I.~Polosukhin, ``Attention is all you need,'' in
  \emph{Advances in Neural Information Processing Systems 30: Annual Conference
  on Neural Information Processing Systems 2017, 4-9 December 2017, Long Beach,
  CA, {USA}}, 2017, pp. 5998--6008.

\bibitem{bottom}
S.~Gehrmann, Y.~Deng, and A.~M. Rush, ``Bottom-up abstractive summarization,''
  in \emph{Proceedings of the 2018 Conference on Empirical Methods in Natural
  Language Processing, Brussels, Belgium, October 31 - November 4, 2018}, 2018,
  pp. 4098--4109.

\bibitem{Bert}
J.~Devlin, M.-W. Chang, K.~Lee, and K.~Toutanova, ``Bert: Pre-training of deep
  bidirectional transformers for language understanding,'' \emph{arXiv preprint
  arXiv:1810.04805}, 2018.

\bibitem{RNN1}
S.~Chopra, M.~Auli, and A.~M. Rush, ``Abstractive sentence summarization with
  attentive recurrent neural networks,'' in \emph{{NAACL} {HLT} 2016, The 2016
  Conference of the North American Chapter of the Association for Computational
  Linguistics: Human Language Technologies, San Diego California, USA, June
  12-17, 2016}, 2016, pp. 93--98.

\bibitem{RNN2}
R.~Nallapati, B.~Zhou, C.~dos Santos, {\c{C}}.~GuÌ‡l{\c{c}}ehre, and
  B.~Xiang, ``Abstractive text summarization using sequence-to-sequence {RNN}s
  and beyond,'' in \emph{Proceedings of The 20th {SIGNLL} Conference on
  Computational Natural Language Learning}.\hskip 1em plus 0.5em minus
  0.4em\relax Berlin, Germany: Association for Computational Linguistics, Aug.
  2016, pp. 280--290.

\bibitem{RNN3}
P.~Li, W.~Lam, L.~Bing, and Z.~Wang, ``Deep recurrent generative decoder for
  abstractive text summarization,'' in \emph{Proceedings of the 2017 Conference
  on Empirical Methods in Natural Language Processing, {EMNLP} 2017,
  Copenhagen, Denmark, September 9-11, 2017}, 2017, pp. 2091--2100.

\bibitem{cnn2}
J.~Gehring, M.~Auli, D.~Grangier, D.~Yarats, and Y.~Dauphin, ``Convolutional
  sequence to sequence learning,'' in \emph{Proceedings of the 34th
  International Conference on Machine Learning}, 05 2017, pp. 1243--1252.

\bibitem{atten}
D.~Bahdanau, K.~Cho, and Y.~Bengio, ``Neural machine translation by jointly
  learning to align and translate,'' in \emph{3rd International Conference on
  Learning Representations, {ICLR} 2015, San Diego, CA, USA, May 7-9, 2015,
  Conference Track Proceedings}, 2015.

\bibitem{atten2}
A.~{\c{C}}elikyilmaz, A.~Bosselut, X.~He, and Y.~Choi, ``Deep communicating
  agents for abstractive summarization,'' in \emph{Proceedings of the 2018
  Conference of the North American Chapter of the Association for Computational
  Linguistics: Human Language Technologies, {NAACL-HLT} 2018, New Orleans,
  Louisiana, USA, June 1-6, 2018, Volume 1 (Long Papers)}, 2018, pp.
  1662--1675.

\bibitem{atten3}
A.~Cohan, F.~Dernoncourt, D.~S. Kim, T.~Bui, S.~Kim, W.~Chang, and N.~Goharian,
  ``A discourse-aware attention model for abstractive summarization of long
  documents,'' in \emph{Proceedings of the 2018 Conference of the North
  American Chapter of the Association for Computational Linguistics: Human
  Language Technologies, NAACL-HLT, New Orleans, Louisiana, USA, June 1-6,
  2018, Volume 2 (Short Papers)}, 2018, pp. 615--621.

\bibitem{template1}
K.~Wang, X.~Quan, and R.~Wang, ``Biset: Bi-directional selective encoding with
  template for abstractive summarization,'' in \emph{Proceedings of the 57th
  Conference of the Association for Computational Linguistics, {ACL} 2019,
  Florence, Italy, July 28- August 2, 2019, Volume 1: Long Papers}, 2019, pp.
  2153--2162.

\bibitem{template2}
Z.~Cao, W.~Li, S.~Li, and F.~Wei, ``Retrieve, rerank and rewrite: Soft template
  based neural summarization,'' in \emph{Proceedings of the 56th Annual Meeting
  of the Association for Computational Linguistics, {ACL} 2018, Melbourne,
  Australia, July 15-20, 2018, Volume 1: Long Papers}, 2018, pp. 152--161.

\bibitem{structure}
K.~Song, L.~Zhao, and F.~Liu, ``Structure-infused copy mechanisms for
  abstractive summarization,'' in \emph{Proceedings of the 27th International
  Conference on Computational Linguistics, {COLING} 2018, Santa Fe, New Mexico,
  USA, August 20-26, 2018}, 2018, pp. 1717--1729.

\bibitem{keyword1}
Q.~Wang and J.~Ren, ``Abstractive summarization with keyword and generated word
  attention,'' in \emph{International Joint Conference on Neural Networks,
  {IJCNN} 2019 Budapest, Hungary, July 14-19, 2019}, 2019, pp. 1--8.

\bibitem{keyword2}
Z.~Deng, F.~Ma, R.~Lan, W.~Huang, and X.~Luo, ``A two-stage chinese text
  summarization algorithm using keyword information and adversarial learning,''
  \emph{Neurocomputing}, 2020.

\bibitem{granularity1}
P.~X. Nguyen and S.~R. Joty, ``Phrase-based attentions,'' \emph{CoRR}, vol.
  abs/1810.03444, 2018.

\bibitem{granularity2}
J.~Hao, X.~Wang, S.~Shi, J.~Zhang, and Z.~Tu, ``Multi-granularity
  self-attention for neural machine translation,'' in \emph{Proceedings of the
  2019 Conference on Empirical Methods in Natural Language Processing and the
  9th International Joint Conference on Natural Language Processing,
  {EMNLP-IJCNLP} 2019, Hong Kong, China, November 3-7, 2019}, 2019, pp.
  887--897.

\bibitem{subword2}
T.~Srinivasan, R.~Sanabria, and F.~Metze, ``Multitask learning for different
  subword segmentations in neural machine translation,'' \emph{CoRR}, vol.
  abs/1910.12368, 2019.

\bibitem{beam}
F.~J. Och and H.~Ney, ``The alignment template approach to statistical machine
  translation,'' \emph{Computational linguistics}, vol.~30, no.~4, pp.
  417--449, 2004.

\bibitem{rouge}
C.-Y. Lin and E.~Hovy, ``Automatic evaluation of summaries using n-gram
  co-occurrence statistics,'' in \emph{Proceedings of the 2003 Human Language
  Technology Conference of the North American Chapter of the Association for
  Computational Linguistics}, 2003, pp. 150--157.

\bibitem{pkuseg}
R.~Luo, J.~Xu, Y.~Zhang, X.~Ren, and X.~Sun, ``Pkuseg: A toolkit for
  multi-domain chinese word segmentation.'' \emph{CoRR}, vol. abs/1906.11455,
  2019.

\bibitem{hugging}
T.~Wolf, L.~Debut, V.~Sanh, J.~Chaumond, C.~Delangue, A.~Moi, P.~Cistac,
  T.~Rault, R.~Louf, M.~Funtowicz, and J.~Brew, ``Huggingface's transformers:
  State-of-the-art natural language processing,'' \emph{ArXiv}, vol.
  abs/1910.03771, 2019.

\bibitem{tf-idf}
P.~Sun, L.~Wang, and Q.~Xia, ``The keyword extraction of chinese medical web
  page based on {WF-TF-IDF} algorithm,'' in \emph{2017 International Conference
  on Cyber-Enabled Distributed Computing and Knowledge Discovery, CyberC 2017,
  Nanjing, China, October 12-14, 2017}, 2017, pp. 193--198.

\end{thebibliography}

%






\vfill



\end{CJK}
\end{document}